\documentclass{article}

\usepackage{arxiv}
\usepackage[utf8x]{inputenc}% allow utf-8 input
\usepackage[T1]{fontenc}    % use 8-bit T1 fonts
\usepackage{url}            % simple URL typesetting
\usepackage{booktabs}       % professional-quality tables
\usepackage{amsfonts}       % blackboard math symbols
\usepackage{nicefrac}       % compact symbols for 1/2, etc.
\usepackage{microtype}      % microtypography
\usepackage{graphicx}
\usepackage{doi}
\usepackage{hhline}
\usepackage{bibentry}
\usepackage{blindtext}

\title{A Survey on GPT-3}

\author{
	Mingyu Zong, Bhaskar Krishnamachari\\
    USC Viterbi School of Engineering\\
    Los Angeles, California 90089\\
    \{mzong, bkrishna\}@usc.edu
}

\hypersetup{
pdftitle={A Survey on GPT-3},
pdfsubject={q-bio.NC, q-bio.QM},
pdfauthor={David S.~Hippocampus, Elias D.~Striatum},
pdfkeywords={First keyword, Second keyword, More},
}

\begin{document}
\maketitle

\begin{abstract}
    This paper provides an introductory survey to GPT-3. We cover some of the historical development behind this technology, some of the key features of GPT-3, and discuss the machine learning model and the datasets used. We survey both academic and commercial efforts applying GPT-3 in diverse domains such as developing conversational AI chatbots, software development, creative work, domain knowledge, and business productivity. We discuss some of the challenges that GPT-3 faces such as the problems of training complexity, bias, and hallucination/incorrect answers. We also discuss the future research opportunities in this area.
\end{abstract}

% keywords can be removed

% keywords can be removed
% \keywords{First keyword \and Second keyword \and More}

\section{Introduction}

\subsection{What is GPT-3?}

GPT-3 is the third generation of the Generative Pre-trained Transformer (GPT) model series, developed by researchers at OpenAI. 

First generation GPT-1 model (also referred as GPT) was built in 2018. It is created to overcome the shortage of labeled data and take advantage of the abundant amount of unlabeled, diverse instances. This auto-regressive decoder-only transformer adopted the self-attention mechanism. It was first trained on the BooksCorpus and then fine-tuned for downstream tasks. In order to test the model's capability, 12 test tasks were selected from 4 major Natural Language Processing (NLP) catrgories: natural language inference, question answering and common sense reasoning, semantic similarity, and classification. This task-agnostic model outperformed state-of-the-art models in 9 test datasets, pointing out another direction towards Artificial General Intelligence (AGI) \cite{radford2018improving}. Under the inspiration of GPT-1's achievements, OpenAI researchers continued using the decoder-only architecture to develop more powerful models. In 2019, they released GPT-2, which had ten times the parameters compared to the first version of GPT and has a pre-training corpus that is ten times larger than its predecessor's. This second model was trained on the WebText, which is a collection of millions of webpages. Having zero supervised training, the enlarged model created new state-of-the-art records in 7 out of 8 language modeling tasks with a zero-shot learning setting. Even though this language model failed to compete with task-specific supervised models on summarization, translation, and other NLP subsets with zero-shot learning, it set convincing baselines that promoted further investigation on fine-tuning the model \cite{Radford2019LanguageMA}.

GPT-3 was published one year later, in 2020. Besides augmenting the model size to contain 175 billion parameters, developers also spent efforts filtering for a high-quality training dataset. As a result, GPT-3 was trained on around 500 billion tokens gathered from five major data sources: Common Crawl, WebText2, Books2, Books2, and Wikipedia ~\cite{brown2020language}.
As a comprehensive commercial product, GPT-3 gets embedded in dozens of real-world applications to offer assistance, including customer service chatbots, report summarization applications, software development tools, etc. More details are provided in section 3. Moreover, OpenAI has made the model available to the public through a playground (https://beta.openai.com/playground) and an API. Every registered user has 18-dollar-worth free credits to try out the model's capabilities. Many parameters, such as mode and temperature, can be adjusted to better fit downstream tasks. Their API gives access to all functionalities of the model provided by the playground, along with an additional fine-tuning option.

\subsection{How to use GPT-3}
\begin{figure}[htp]
    \centering
    \includegraphics[width=17cm]{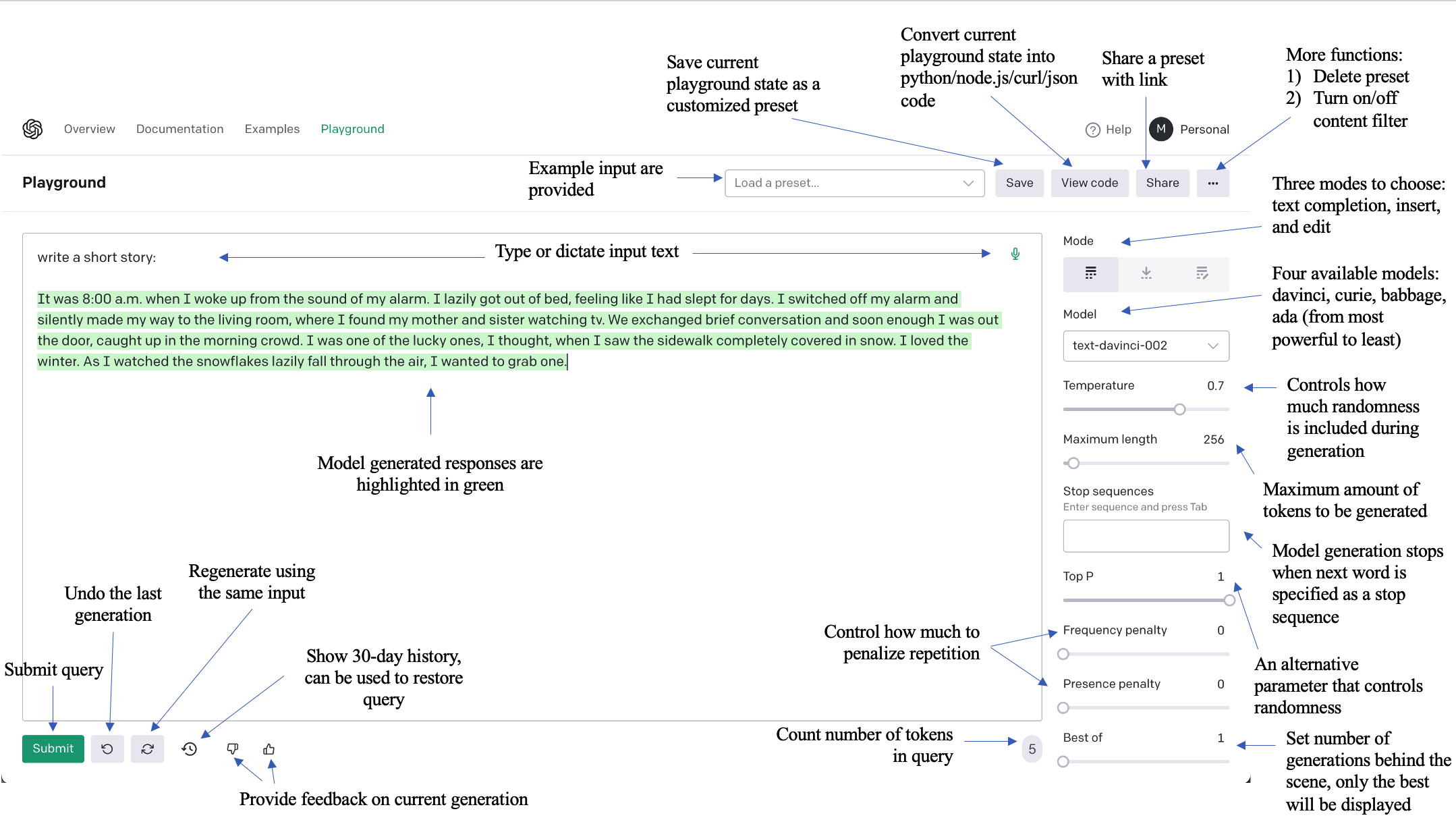}
    \caption{GPT-3 Playground under the "completion" mode}
    \label{fig:playground}
\end{figure}

The GPT-3 playground is a published website with explicit design to help users get start using the model. Users can choose between 4 models and 3 different modes based on their needs. The most frequently used mode is the "complete" mode. A user input some text and expect the model to continue writing until the next word or sequence to be generated is a stop sign. The "insert" mode requires users to use the placeholder "[insert]" when submitting a query. GPT-3 will return the whole piece of text with the placeholder being substituted by its prediction. However, the model can deal with only one insert at a time. To use the "edit" mode, input text and instructions must be typed into two text boxes. Output of the model would be modified input text. A typical usage of the edit mode is grammar correction. The 4 models include davinci, curie, babbage, and ada. Cost of using a model is proportional to model's ability, so customers can pick the optimal one to use base on their needs and budget. Among these, davinci is the most powerful yet most expensive model. The remaining three models trade off performance for being more economical. Nevertheless, ada, the least capable model, is sufficient to handle some simple tasks.

A set of parameters are listed on the right hand side of the input text box, under the selection of mode and model. By adjusting values of these parameters, users are able to add some control over the model's behavior. One of the most frequently used parameters is the "temperature", which determines how much randomness to involve during generation. Generative models are well known for their ability to generate new contents with the same input prompt \cite{gero2022we}. GPT-3 achieves this through the "temperature" parameter. If the value is set to 0, then regardless of number of times of generation, the model will always output the same response with fixed input text. Increasing the value of "temperature" allows users to take advantage of GPT-3's creativity, it will provide more possible completions from multiple runs. Maximum value allowed for this is 1. "Maximum length" sets the upper limit of the amount of tokens to be generated. Default value is 256, but writing paragraphs or stories often requires more. A maximum of 4000 tokens can be output. "Stop sequence" is another useful parameter. If the model is expected to play a role in a conversation, adding a delimiter is necessary to prevent it from writing the lines for the other character. Other available parameters are "TOP P", "Frequency penalty", "Presence penalty", "Best of", "Inject start text", "Inject restart text", and "Show probabilities". All of these can be adjusted under the "completion" mode. However, both the "insert" and the "edit" modes are in beta, so there are less available options provided.

Furthermore, people can interact with GPT-3 through OpenAI's API. Before use, the "openai" library must be installed and a user needs to acquire an API key. Then the user can access the model with specified input, model, mode, and other parameter settings to use GPT-3 in one of the three modes listed before. One function that provided by the API but not the playground is fine-tuning, which further trains the model to adapt to specific task(s). Since it is allowed to keep training a fine-tuned version of GPT-3, with enough resources, the model's overall ability can also be lifted. Prerequisite of using the fine-tuning functionality, is a prepared JSONL file, in which every instance is broken into two parts: "prompt" and "completion". OpenAI has a corresponding CLI data preparation tool, which easily converts a CSV, TSV, XLSX, JSON or JSONL file into the required format. Besides model, other parameters such as "n\_epochs" and "batch\_size" can be customized. After completion of the fine-tuning, the model will be accessible by its unique name. GPT-3 learns patterns from training instances in a provided JSONL file. When the fine-tuned model sees a similar test instance, it uses new knowledge to write better answer.

\subsection{Model performance on NLP tasks}
\begin{table*}[ht]
\fontsize{6.5}{6.5}\selectfont
\centering
\def\arraystretch{1.5}%
\begin{tabular}{ |c c c c c| }
\hline
 & & & \multicolumn{2}{c|}{\textbf{175B GPT-3}}\\
\hhline{}
\textbf{Dataset} & \textbf{Metric} & \textbf{Fine-tuned SOTA} & \textbf{Zero-shot} & \textbf{Few-shot}\\
\hline
\multicolumn{5}{|l|}{Language Modeling}\\
\hline
PTB & perplexity & 35.8 & \textbf{20.5} & - \\
\hhline{}
 & & & &\\
\hhline{}
\multicolumn{5}{|l|}{Text Completion}\\
\hline
LAMBADA & accuracy & 68.0 & 76.2 & \textbf{86.4} \\
StoryCloze & accuracy & \textbf{91.8} & 83.2 & 87.7 \\
HellaSwag & accuracy & \textbf{85.6} & 78.9 & 79.3 \\
\hhline{}
 & & & &\\
\hhline{}
\multicolumn{5}{|l|}{Question-Answering}\\
\hline
NaturalQS & accuracy & \textbf{44.5} & 14.6 & 29.9 \\
WebQS & accuracy & \textbf{45.5} & 14.4 & 41.5 \\
TriviaQA & accuracy & 68.0 & 64.3 & \textbf{71.2} \\
\hhline{}
 & & & &\\
\hhline{}
\multicolumn{5}{|l|}{Language Translation}\\
\hline
En -> Fr & BLEU-mb & \textbf{45.6} & 25.2 & 32.6 \\
Fr -> En & BLEU-mb & 35.0 & 21.2 & \textbf{39.2} \\
En -> De & BLEU-mb & \textbf{41.2} & 24.6 & 29.7 \\
De -> En & BLEU-mb & 40.2 & 27.2 & \textbf{40.6} \\
En -> Ro & BLEU-mb & \textbf{38.5} & 14.1 & 21.0 \\
Ro -> En & BLEU-mb & \textbf{39.9} & 19.9 & 39.5 \\
\hhline{}
 & & & &\\
\hhline{}
\multicolumn{5}{|l|}{Understanding Reference of Pronoun}\\
\hline
Winograd & accuracy & \textbf{90.1} & 88.3 & 88.6 \\
Winogrande(XL) & accuracy & \textbf{84.6} & 70.2 & 77.7 \\
\hhline{}
 & & & &\\
\hhline{}
\multicolumn{5}{|l|}{Commonsense Reasoning}\\
\hline
PIQA & accuracy & 77.1 & 81.0 & \textbf{82.8} \\
ARC(easy) & accuracy & \textbf{92.0} & 68.8 & 70.1 \\
ARC(challenge) & accuracy & \textbf{78.5} & 51.4 & 51.5 \\
OpenBookQA & accuracy & \textbf{87.2} & 57.6 & 65.4 \\
\hhline{}
 & & & &\\
\hhline{}
\multicolumn{5}{|l|}{Reading Comprehension}\\
\hline
CoQA & F1 & \textbf{90.7} & 81.5 & 85.0 \\
SQuADv2 & F1 & \textbf{93.0} & 59.5 & 69.8 \\
RACE-h & accuracy & \textbf{90.0} & 45.5 & 46.8 \\
RACE-m & accuracy & \textbf{93.1} & 58.4 & 58.1 \\
\hhline{}
 & & & &\\
\hhline{}
\multicolumn{5}{|l|}{SuperGLUE Benchmark}\\
\hline
SuperGLUE & average & \textbf{89.0} & 58.2 & 71.8 \\
\hhline{}
 & & & &\\
\hhline{}
\multicolumn{5}{|l|}{Understanding Relationship between Sentences}\\
\hline
ANLI R1 & accuracy & \textbf{73.8} & 34.6 & 36.8 \\
ANLI R2 & accuracy & \textbf{50.7} & 35.4 & 34.0 \\
ANLI R3 & accuracy & \textbf{48.3} & 34.5 & 40.2\\
\hhline{}
 & & & &\\
\hhline{}
\multicolumn{5}{|l|}{Arithmetic}\\
\hline
2 digit addition & accuracy & - & 76.9 & 100.0 \\
2 digit subtraction & accuracy & - & 58.0 & 98.9 \\
3 digit addition & accuracy & - & 34.2 & 80.4 \\
3 digit subtraction & accuracy & - & 48.3 & 94.2 \\
4 digit addition & accuracy & - & 4.00 & 25.5 \\
4 digit subtraction & accuracy & - & 7.50 & 26.8 \\
5 digit addition & accuracy & - & 0.65 & 9.30 \\
5 digit subtraction & accuracy & - & 0.80 & 9.90 \\
2 digit multiplication & accuracy & - & 19.8 & 29.2 \\
1 digit composite operations & accuracy & - & 9.75 & 21.3 \\
\hhline{}
 & & & &\\
\hhline{}
\multicolumn{5}{|l|}{Manipulated Words Correction}\\
\hline
cycle letters & accuracy & - & 3.66 & 37.9 \\
anagrams 1 & accuracy & - & 2.28 & 15.1 \\
anagrams 2 & accuracy & - & 8.91 & 39.7 \\
random insertion & accuracy & - & 8.26 & 67.2 \\
\hhline{}
\multicolumn{5}{|l|}{SAT Analogies}\\
\hline
SAT Analogies & accuracy & - & 53.7 & 65.2 \\
\hline
\end{tabular}
\caption{Various Test Results From SOTA And GPT-3}
\label{table:GPT-3 results}
\end{table*}

Along with the publication of the model, OpenAI developers also performed thorough examination on GPT-3 of all sizes and presented results in their original work~\cite{brown2020language}. Every model was tested three times using zero-shot, one-shot, and few-shot learning methods separately. Zero-show learning means no demonstration is provided to a model; while few-shot learning will offer a few examples for a model to learn, then ask for a completion. Observing from the results, model performance is expected to improve as a) model size increases or b) more demonstrations are available. We present some of the results that Brown \emph{et al.} obtained from full-sized (175B) GPT-3 and benchmarks set by SOTA models for reference (Table~\ref{table:GPT-3 results}), more details can be found in the original paper\cite{brown2020language}.

As a language model, GPT-3 is as competitive as the fine-tuned SOTA model on several aspects, including text completion, question-answering, translating other language into English, and understanding reference of pronoun. Among all these, it is most capable of text completion and determining reference, with an overall accuracy above 80\% on these tasks. On the other hand, its performance on the rest three categories are remarkable but not yet satisfying as an alternative of human intelligence. One variation not shown in the table is that, for question-answering test, fine-tuned SOTA model is open domain, but GPT-3 had no access to any external knowledge base during the process. Under these circumstances, GPT-3 is comparable as SOTA in two of the three cases. Based on the considerable gaps between zero-shot results and few-shot results, researchers suggest that problems in the NaturalQS dataset and the WebQS dataset could be out of distribution for GPT-3's training set~\cite{brown2020language}. Improvement of future language model can be built on this to fully explore the potential of transformer. More complicated assignments such as commonsense reasoning and reading comprehension still remain challenging for this architecture, even though the "attention" mechanism enhanced its comprehension on input text: it can hardly succeed with more than 70\% accuracy. Besides ordinary NLP tasks, GPT-3 was also tested for its ability in doing arithmetic, correcting misspelled English words, and solving ``SAT Analogies". Most of the returning success rates are below 50\%. But the model displays distinctive strength in arithmetic computation when operands have at most three digits~\cite{brown2020language}.

Decoder-only transformer models usually perform better at language generation. Developers of GPT-3 examined this capacity through article generations and human evaluation. As a result, human successfully distinguish AI-generated passages from authentic ones at a rate of 52\%, which is merely above chance~\cite{brown2020language}.

\section{Historical Background and Prior Literature before GPT-3}

\subsection{Early NLP Models}
Human efforts in solving natural language processing (NLP) can be traced back to the 1950s, as translation between languages appeared to be the key to intelligence collection during World War II. Linguisticians tried to decode the foreign symbols and summarized a set of rules to follow for translation. Early NLP experiments programmed the machine to automatically enforce those rules when performing translation tasks, contributing to one of the most typical assignments in NLP: machine translation (MT). However, human assistance was often involved due to the limited coverage of foreign language regularities, which contradicts the idea of a fully automized process. An expert editor would revise a preliminary translation result before it could be put to use. Researchers also utilized multiple iterations of human revision and re-generation with updated dictionaries and grammar to optimize the quality of instance generation and the whole translation system ~\cite{Hutchins1995MachineTA}. The Automatic Language Processing Advisory Committee released a report in 1966 that thoroughly assessed MT systems' development. Given the numerous resources invested into the field of study, the performance of most advanced systems at the time with human post-editing still performed much worse than human translators regarding accuracy, efficiency, and comprehension level; conclusively, the committee claimed there was no immediate prospect of a desirable system ~\cite{Pierce1966LanguageAM}. Machine translation research reached a trough under the reflection until statistical MT systems emerged to replace symbolic systems. Starting with IBM's promising experiments of purely statistical methods, subsequent research focused on constructing corpora and adopting machine learning methods to align words and phrases between languages ~\cite{Schwartz2018TheHA}. With increasing computational power and evolving designs of processing architecture, natural language processing extended its scope to include text summarization, document classification, question answering, and more sub-classes.

Before the introduction of Transformer models, Recurrent Neural Network (RNN) models gained significant popularity in solving NLP tasks. These models have the capability to memorize and process long-term dependencies among inputs because of the additional hidden layer embedded in the architecture. Thus, they appeared to be more effective at understanding sequential and time series data. This nature has led to its prominent manifestation in healthcare. A case study of clinical diagnoses and medication group prediction using EHR records has proved that the RNN model performed better with longer input records ~\cite{Xiao2021RecurrentNN}. RNN models' convincing ability to complete classification assignments has also benefited other fields, such as human studies and city planning. Nevertheless, the design of these models resulted in a slow computation rate, making it challenging to train an RNN model over a large dataset. Consequently, not being able to exploit the value of the enormous amount of data is why scientists developed the superseding Transformer architectures.

\subsection{Transformers Before GPT-3}

In 2017, the Google Brain research team proposed a "self-attention" mechanism, along with the vanilla Transformer network architecture built upon it. The architecture inherits the encoder-decoder structure. However, instead of highlighting relationships between sequences, self-attention captures relationships between different parts within one sequence by packing key-value pairs of vectors (i.e., representations of pairwise relationships) into the computation matrix ~\cite{Vaswani2017AttentionIA}. Lin pointed out in the survey that the encoder or decoder alone is sufficient for solving specific types of NLP problems ~\cite{Lin2021ASO}. Generative Pre-training Transformers (GPT) is the first published model pre-trained on a large corpus; the model is decoder-only. Developers of GPT at OpenAI tested the process of combining pre-training and fine-tuning and proved it could boost model performance drastically. This workflow ensures that GPT is adaptive to multiple tasks while overcoming the difficulty of not having enough labeled data ~\cite{radford2018improving}. A series of pre-trained Transformer models were created in recent years, with some variations in design. Bidirectional Encoder Representations from Transformers (BERT) with the complete encoder-decoder architecture is one of the most prominent. Its bi-directionality allows the model to relate both previous and following words when attending to any individual within the input ~\cite{Devlin2019BERTPO}. As transformer models enlarge their scales, the community has agreed that model performance is proportional to its scale. Therefore, top tech companies are investing resources into creating a more powerful Transformer to explore its upper limit.

\section{(Non-Academic) Applications of GPT-3 in Real World}
\subsection{Assistance in Question-Answering System}
There are mainly two types of products that exploits GPT-3 for assistance in question-answering interaction. On is customer service providers, and the other is chatbots. Customer service sometimes is delivered through a chatbot. For example, \href{https://activechat.ai/} {ActiveChat} allows users to chat with their customers at real time. At the same time, GPT-3 offers hints to help users formulate their responses. On the other hand, customer service can be offered by replying to customers' comments/feedback/reviews. \href{https://www.replier.ai/} {Replier.ai} free retailers from manually responding to customers' reviews. They offer a paid product which automatically replies to the reviews after simple setup. Chatbots are developed for various purposes, ranging from entertainment to simulation of the dead. \href{https://aibuddy.chat/} {AIBuddy} is a GPT-3 powered version of Whatsapp. A user can discuss whatever topic he/she is interested in with the agent. \href{https://projectdecember.net/} {Project December} is a hyper-realistic chatbot to simulate conversations with any human being, 
alive or not.

\subsection{Contribution to Creative Work}
Many applications on the market are powered by GPT-3 to generate creative and customized contents to meet users' demand. For instance, \href{https://copysmith.ai/} {Copysmith} combined GPT-3 with other machine learning algorithms to write texts based on users' requirements. Description of product, advertisement headline, blog idea are all covered by its functionality. Similarly, \href{https://encharge.io/free-ai-email-subject-line-generator/} {email subject line generator by Encharge} utilizes the model to write emails. Required input to use this application is a single-line subject description. And users can choose between more than ten tones for a desired writing style. \href{https://www.botto.com/} {Botto}, another software, uses GPT-3 in a different way. Input data is supposed to be an image, and the model generates poetic description for it. Another popular usage of the model is writing articles and essays. \href{https://culture.org/ghosts/} {Ghosts} is a special app that writes essay about grief with support from GPT-3.

\subsection{As Software Developer}
The \href{https://themesberg.com/blog/tailwind-css/gpt-3-tailwind-css-ai-code-generator} {Tailwind CSS code generator} is one of the code generation applications that empowered by GPT-3 model. It is design to write CSS code based on requirements in textual format. However, later in 2021, OpenAI developers created the Codex system based on GPT-3, which is specifically designed to convert natural language description into code. More code generation applications chose to embed Codex instead of the original GPT-3 model because of the greater ability in this specific task. \href{http://ww1.codestart.xyz/} {CodeStarter} is a representative application that uses Codex to assist code generation. With a description of intended web app, CodeStarter will write the scripts instantly in any specifies framework: React, Angular, and more.

\subsection{Additional Usage by Domain}
Besides the above three common category, GPT-3 is also used in games, language learning, and content searching. Specifically, \href{https://aidungeon.io/} {AI Dungeon} relies on GPT-3 to generate a new set of exploration options for a text adventure game. \href{https://www.duolingo.com/} {Duolingo} uses GPT-3 to correct grammars of French sentences to ensure the materials won't mislead its users. \href{https://www.podacity.ai/} {Podacity AI} trusts GPT-3's ability in understanding natural language. So the model is in charge of understanding what kind of podcasts are wanted by the user.

\section{Academic Research on GPT-3 and its Applications}

\subsection{GPT-3 for Bio-medical and Healthcare domain}
One classic usage that makes GPT-3 helpful in healthcare is that it can be trained to support customer service with minimum efforts. Image having an automatic agent instantly replying to trivial questions and translate local languages into English, resources at a clinic can be more efficiently allocated \cite{Nath2022NewMF}. Beyond these, it is more difficult but plausible to train the model for support in triaging patients. GPT-3 can acquire all necessary information through a conversation instead of a intake form, and there could be multiple implemented systems to help a number of patients simultaneously \cite{Korngiebel2021ConsideringTP}. Overall, efficiency is the most attractive reason of adopting a language model in this area. There are other factors at play as well. Clinics and hospitals are places to cure illness, but it is also true that various contagious viruses are brought into these places. If a human receptionist is replaced by a robot, both patient and healthcare provider can be better protected from being infected. This consideration appears to be even more essential during the Covid-19 pandemic.

Some major concerns of embedding this language model into healthcare delivery process include cost, model behavior, and privacy concerns. Clinics and hospitals usually have a high volume of daily visits. An optimal system may submit multiple queries for every single patient in order to generate appropriate output. Altogether, deployment of GPT-3 could result in considerable amount of operation cost. Furthermore, existing auditing methods are not mature enough to suppress all problematic generations, as we further explore in section 4.4. Healthcare is a field with even higher standards for interactions with patients. Thus, it is worth evaluating the risks and harms before implementation. Finally, medical records are strictly protected by convention. But seeking help from GPT-3 would inevitably share these data with OpenAI. It is necessary that OpenAI and healthcare providers obey the same set of rules to protect patients' privacy \cite{Sezgin2022OperationalizingAI}.

\subsection{GPT-3 for writing}
Generative language models are prominent for writing authentic texts in different format, including articles, essays, and summaries. In one study, a one-sentence prompt is offered to the GPT-3 and human writers for continuation of one paragraph. And the generated texts are randomly sent to human annotators for evaluations. Researchers quantified different error types noticed in the generations. It turns out that GPT-3 is better at using technical jargon and knowledge that human readers often need to google to confirm. But human writers outperformed the model with respect to the other eight error types. Another finding from the study is that model's overall ability in writing is growing with model's size \cite{Dou2022IsGT}. Apparently, contents generated by GPT-3 are still distinguishable from human writings, but it's possible that newer language models are able to produce more reliable sources and eventually defeat human in open-ended composition. Although having GPT-3, or language model in general, as a writer raises problems with copyrights, which we investigate in section 5.2. Another recent work aims to use GPT-3 at data augmentation. By using ensemble techniques and 210 example dialog summaries, Chintagunta was able to obtain medically correct synthesized summaries \cite{Chintagunta2021MedicallyAG}. It is known that medical records are not easily accessible because of the regulations. This project proposes a solution for acquiring medical data without violates patients' privacy. Without the challenge of data acquisition, the research community can make faster progress in AI-assisted medical experiments.

\subsection{GPT-3 for generating code or plans}
Although not as reliable as in common NLP tasks like text summarization, GPT-3 shows its potential in writing code blocks and programming scripts. Examining 80 programs generated by the model, researchers found that 38 of them are of good quality. A conclusion is derived from the test results that GPT-3's code generation can partially contribute to software development \cite{Narasimhan2021CGEMsAM}. It is still too early to replace human coders with an AI agent.

Another area that GPT-3 expanded to cover is plan formulation, which primarily relies on human judgment from natural language description of the situation. With large language models' natural language understanding talent, researchers see a possibility to free human resources from repeatedly laying plans for similar tasks. Olmo adopted a few-shot setting (from 1-shot to 4-shot) to test whether GPT-3 is good at plan extraction or not. Microsoft Windows
Help and Support, WikiHow Home and Gar-
den and CookingTutorial datasets were included in the experiment. Among all four models, davinci outperformed the other three with over 80\% F1 score, and its performance benefits from additional examples provided in the prompt. Even though GPT-3 can compete with state-of-the-art methods, the fact that it learns from available examples makes it another task-specific approach. Its problem-solving ability cannot extend to other scenarios \cite{Olmo2021GPT3toplanEP}. Another research project focus on using GPT-3 to write Lesson plans. It deployed a zero-shot learning setting. All prompts contained only a one-sentence instruction. Although some contents in the responses are incorrect, there is an opportunity to utilize GPT-3 in forming K-12 lesson plans \cite{Walsh2022LessonPG}.

\subsection{GPT-3 for math tutoring}
Educators have long been struggling with the imbalanced ratio of education resources to student demand in various disciplines. One of main issues is that there lack accessible teaching materials for students to study off-school. Online tutorial videos are emerging to ease the situation, but society still hope to provide personalized teaching services through AI to meet the learning needs of each student and improve teaching efficiency. Among many disciplines, mathematics is relatively more systematic. Researchers are making progress towards an artificial math tutor. Recurrent neural network (RNN)-based sequence-to-sequence, graph-to-tree, and sequence-to-tree networks are constructed to convert math word problems into representative equations \cite{wang2017deep, zhang2020graph, wu2020knowledge}. Transformer-based architectures are also proposed to take advantage of a huge volume of training data. However, even though these models performed as expected on test set, they fail to extend their ability to convert out-of-scope math problems. Besides, previous studies mainly focus on questions with only one variable. On the other hand, GPT-3 is able to extract a system of equations which is composed of two unknowns with satisfying performance. It achieved an accuracy of 69\% with a few-shot learning setting and further improved the performance to convert 80\% of the questions correctly with fine-tuning\cite{zong2022solving}.

\subsection{Auditing}
Due to the unconstrained output space, it remains challenging to interpret a large language model. The difficulty will be passed on to future models. For it to be used as (part of) a commercial product, developers of GPT-3 are expected to audit the model thoroughly before it's open to the public. However, there is no systematic way to unravel model's problematic behavior. The field is drawing increased attention from the research community.

Some preliminary actions taken by OpenAI researchers include content filtering and feedback collection. An automated content filter is used to constantly screening output of GPT-3 for regulating purpose. Every generation is classified into one of the three levels: safe, sensitive, and unsafe. Depending on the classification result, generated text is either suppressed or displayed with a warning. Documentation of GPT-3 emphasizes that the function is still in beta and results in higher false positive rate. An example that confirms the filter's imperfection is found by Gero. One generated text that contains the word "lesbian", despite the whole statement being positive, induced a warning of unsafe content \cite{gero2022we}. In this case, the harm is not caused by a toxic answer but the design of the filter. Another issue with the filter design is that marginalized identities often don't get enough consideration \cite{Bender2021OnTD}. Collection of user feedback on each generation is another insurance that OpenAI researchers take to monitor and adjust GPT-3's behavior. Users with diverse background help reduce bias in decision-making. However, the process takes time and fails to protect whoever sees a harmful content at the first place. Weighing both pros and cons of the existing methods, it is clear that another efficient and accurate auditing algorithm/system is highly in demand.

\subsection{Research on improving GPT-3 performance}
Primary method to improve GPT-3's performance is \emph{prompt engineering}. This refers to what instructions are provided and whether or not to offer examples in a prompt. More importantly, prompt engineering is not task-agnostic. There is no standardized setting that guarantees the best performance in every test. Due to difficulty of interpretation, it is always possible that a better prompt exists for a specific task. With limited input space, exhaustive testing is a way to find out the best combination of instructions and examples. But this method is computationally expensive. So researchers keep exploring more robust solutions. With respect to example selection, the best practice is to provide similar ones so that it is easier for a model to catch the point. Therefore, Prodan and Pelican developed a scoring system to rank conversations based on similarity. Three concepts are taken into account: length, content, and number of characters involved. Components are multiplied by weights and then summed up to calculate the final score \cite{Prodan2021PromptSS}.

On the other hand, OpenAI researchers trained a verifier, which in fact is a fine-tuned GPT-3, to output possibilities of an input being correct. The entire problem-solving system consists of a generator and a verifier. Both of them are GPT-3 models, but are fine-tuned using different dataset. For every test instance, multiple candidates are pooled from the generator and ranked according to the verifier's evaluation. The one obtained the highest probability is considered the final output \cite{Cobbe2021TrainingVT}. In other words, GPT-3 itself can be calibrated into a selection tool when fine-tuning is appropriate for intended results.

\section{Limitations and Pitfalls of GPT-3}

\subsection{Limitations in Design}
As revealed in OpenAI's testing results, many NLP problems and mathematical calculations expressed in the textual format are still beyond the scope of GPT-3's knowledge. Thus, transformers' performance on several tasks has not yet reached the cap. Since the model's performance on various tasks still grows with scale, researchers can efficiently address this issue by expanding a pre-trained transformer's training dataset and enlarging the model size. Given its unidirectional decoder-only architecture, GPT-3 is naturally less suitable for certain types of NLP tasks, for example, semantic understanding. Using only statistical probabilities to predict the following tokens, GPT-3 tends to fail simple reversible answers that can be solved by common knowledge ~\cite{floridi2020gpt}. Moreover, GPT-3 is primarily designed to process text inputs. Although when using the playground, users can upload audio file or record instantly to have the data converted into text, the functionality is still in beta. Developing extensions to precisely transform imagery and audio input will further expand model's usage in the industry. Lastly, even though GPT-3 has approved its capability in various assignments to excel in domain-specific tasks, fine-tuning is still required to supplement domain knowledge and improve performance. A reasonable amount of data required to fine-tune GPT-3 is 1000 instances, which could be a challenge for some down-stream tasks, for example, analysis of medical records.

\subsection{Misuse of the model}
One concern pointed out by GPT-3 developers is the misuse of this model, including fraudulent essay composition, spam message creation, and even assistance in formulating malicious tactics, techniques, and procedures (TTPs). Given the fact that GPT-3 is writing on top of its training corpus, it is difficult to decide who should get credit for the generated piece of work: writers who composed these training data, GPT-3's owner OpenAI, or whoever used the model to create the text. This nature also blurs the boundary between plagiarism and original work \cite{Dehouche2021PlagiarismIT}. The model currently couldn't support any large-scale activities because generated contents still need to be manually filtered for quality \cite{brown2020language}. However, GPT-3 is not embedded with mechanisms to detect and prevent illegal usage, which may lead to the development of subsidiary systems and new regulations or policies pertaining to authentic materials and plagiarism.

\subsection{Bias}
Humans are inevitably biased, so are their writings. Being trained on human created contents infers that GPT-3 inherits these biases and will perpetuate them using its responses. Several studies focused on eliciting biased generations. Different types of biases are revealed. First type of bias relates to gender representation. GPT-3's output space varies depending on the perceived gender of a given character. Four topics that are frequently used to generate stories for a male character are war, crime, politics, and sports. However for a female character, her story depicts more about family, emotions, and body. Furthermore, masculine descriptions are often used to portray a male character, while female characters generally have a weaker and less powerful setting \cite{Lucy2021GenderAR}. Even if the same set of brilliant traits are offered in the prompt, the model still understates the brilliance of woman compared with a man \cite{Shihadeh2022BrillianceBI}. Second variation of bias that exists in GPT-3's "mind" is racial bias. OpenAI researchers explored whether racial bias would be a major concern for publication. A placeholder of race is embedded in every prompt and the model is supposed to choose among seven available options. After computing sentiment score, they noticed that sentiment score for "Asian" is consistently high but the score for "Black" is consistently low \cite{brown2020language}. Researchers at Stanford University created a Q-Pain dataset to test if GPT-2 and GPT-3 has racial biases in medical Q\&A tasks. The experiment focused on pain management. As a result, GPT-3 has 3.6\% higher possibility of refusing a pain treatment if the patient is black \cite{Loge2021QPainAQ}. This set of discoveries support that GPT-3 is racially biased and it needs further calibration. Moreover, religious bias is also an issue with large language models including GPT-3. A research found out that the word "Muslim" is tightly related to the word "terrorist" and has a much higher chance to induce violent output \cite{Abid2021PersistentAB}.

\subsection{Being energy-intensive}
Running such a gigantic architecture is incredibly energy-consuming. In addition, the training process is equally energy-intensive, considering the resources needed to distribute billions of tokens and number of iterations required before publication \cite{Bommasani2021OnTO}. Future language models aim to surpass GPT-3 by increasing model size until another architecture is proposed to replace transformers. There is no doubt that resources to be consumed will grow exponentially, leading to serious environmental concerns. It is worth thinking whether the boost in model performance deserves the extra amount of resources needed for its operation. Considering the accelerated global warming and its negative impact on human habitat, it is imperative to innovate new architecture which is equivalently powerful yet more environmental-friendly.

\section{Conclusion}
We have introduced the GPT-3 model, arguably the most capable large language model in existence today. This introduction discussed what it is, how it is developed, what it's capable of, and what its limitations are. As the largest publicly available language model at this time, GPT-3 performs well in a number of standard NLP tasks such as text completion and summarization. Despite the risks of creating misleading/inappropriate/biased, this architecture is already starting to see practical adoption in commercial products, from chatbots to software development applications. However, the research community emphasizes the importance of development of auditing systems, which will benefit not only GPT-3, but also future language models. 

\bibliographystyle{alpha}
\bibliography{references}

\end{document}